# Exploratory Study: Children's with Autism Awareness of Being Imitated by Nao Robot

Andreea Peca[1], Adriana Ţăpuş[2], Amir Aly[2], Cristina Pop[1], Lavinia Jisa[1], Sebastian Pintea[1], Alina Rusu[1], and Daniel David[1]

[1] Babeş-Bolyai University, Department of Clinical Psychology and Psychotherapy, 37 Republicii Street, Cluj-Napoca, Romania

andreea.peca@gmail.com, pop.cristina@ubbcluj.ro, lavinia_jisa@gmail.com, sebastianpintea@psychology.ro, alinarusu@psychology.ro, danieldavid@psychology.ro

[2] Ecole Nationale Superieure de Techniques Avancees (ENSTA-ParisTech), UEI, Cognitive Robotics Lab, 32 Blvd Victor, 75015, Paris, France

adriana.tapus@ensta-paristech.fr, amir.aly@ensta-paristech.fr

**Abstract:**

This paper presents an exploratory study designed for children with Autism Spectrum Disorders (ASD) that investigates children's awareness of being imitated by a robot in a play/game scenario. The Nao robot imitates all the arm movement behaviors of the child in real-time in dyadic and triadic interactions. Different behavioral criteria (i.e., eye gaze, gaze shifting, initiation and imitation of arm movements, smile/laughter) were analyzed based on the video data of the interaction. The results confirm only parts of the research hypothesis. However, these results are promising for the future directions of this work.

**1. Background**

Autism Spectrum Disorders (ASD) refers to a group of conditions characterized by impairments in social interaction, deficits in communication and restricted and repetitive behaviors and/or interests [1]. Existing studies from the literature show that children with ASD have a great affinity with mechanical components, computers, and robots. Early imitation is one of the most important instruments for social learning [6]. There are evidences that preverbal children use imitation as a communication tool and that this function disappear as the verbal language emerges, suggesting that imitation is a prerequisite for the verbal language [5].

Many clinical studies claim that children with autism show a deficit in imitation [8], therefore causing serious consequences for their social interaction and communication. Improving the imitation skills of children with autism through specifically designed treatments based on imitation may yield, on the long term, to an improvement in language, play skills and theory of mind [3], [7].

In [5], Nadel claims that through being imitated, low-functioning children with autism start understanding that they can be at the origin of intentional actions of others. Some clinical studies suggest the effectiveness of adults imitating children with autism and vice versa [4], [2] in play situations by enhancing imitation, recognition of being imitated and nonverbal communication [5].

## 2. Aim/Purpose

Based on these findings, in this study, we investigate if children with autism show more social engagement when interacting with a robot that is mirroring their movements compared to a human partner in a motor imitation game. As an implicit assumption, we believe that the detection of the contingency between the robot's movements and own movements, leading to an awareness of being imitated by the robot, can be the underlying mechanism of such an increase in social behaviors.

The goal of our study is to facilitate the spontaneous imitation and therefore increase the arm motor initiations of children. In this context, we expect that the robot, contingently mirroring the child's arm movements, will act as a natural reward system, increasing the motivation to enhance the rate of initiations and imitations.

## 3. Methods

The study lasted 4 weeks. Each of the 4 children involved had 2 intervention sessions per day: one for the dyadic interaction and one for the triadic interaction, separated by a 10 minutes brake. A single-subject ABAC (A-baseline; B-robot; A-baseline; C-person) design with replication across the 4 participants was employed. Nao robot imitates gross arm movements of the child in real-time in dyadic and triadic interactions. The independent variable is the type of interaction agent: robot versus human person. The dependent variables measured were: frequency of free initiations (without prompt), frequency of total initiations (with and without prompt) and imitations, duration of attention to the partner (robot/person), duration of smile and/or laughter, and frequency of gaze shifting between the two partners in triadic interactions.

The baseline procedures A(1) and A(2) consisted of a motor imitation game between the experimenter and the child. The game has the dynamics of a natural interaction game a child and another partner usually play. The experimenter produced several motor acts, waiting after each behavior for the child to initiate a motor act in response and mirroring this movement when it appeared. If the child did not produce any motor act in a 10 seconds period, the experimenter produced another initiation. The B stage or the robot interaction stage was composed of several conditions: Familiarization, Modeling, Dyadic Interaction, and Triadic Interaction. During the dyadic phase, the robot mirrors each of the following gestures performed by the child: rising left/right arm vertically, rising both arms vertically; rising left/right arm horizontally; rising both arms horizontally; simulating flight (see Figure 1). If the child does not initiate motor acts, the experimenter offers motor prompting. In the triadic phase, the same game is performed, but if the child does not initiate movements, instead of a motor prompting, he is offered a modeling prompting (the experimenter initiates movement). In the C stage or the person interaction stage the procedure was analogue to the procedure in the B stage, but instead of a robot, the child interacts with a human person.

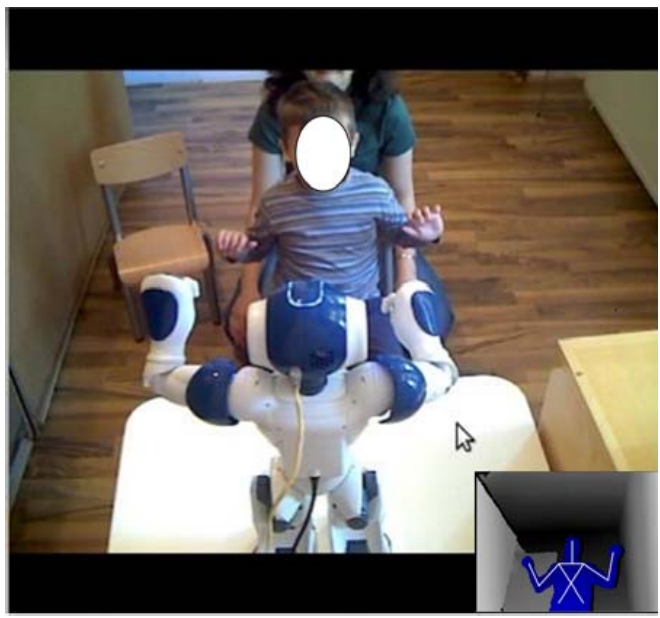

Fig. 1: Child#4 Interacting with Nao Robot and the Corresponding Depth Image of the Operator (the third person staying on the other side of the room).

**4. Environmental Setup**

The study was conducted in a 4m x 4m testing room (see Figure 2) that was split in two areas by a false wall. The left part of the room featured a table and two chairs (one for the child and one for the experimenter). The child interacted directly with the robot that was positioned on the table. In the right part of the room the operator controlled the robot's movements by using the Kinect sensor and by observing the child movements on a computer connected to a webcam. The depth data coming from the Kinect camera is used by the Prime Sense middleware in order to perform skeleton tracking. The skeleton data (i.e., joints and rotation positions) is transmitted to Nao robot who's mirroring the user upper-body position.

We were constraint to use Wizard-of-Oz (WoO) technique and to split the room in two parts, because otherwise it would have been very difficult for the children to stay still for the calibration process.

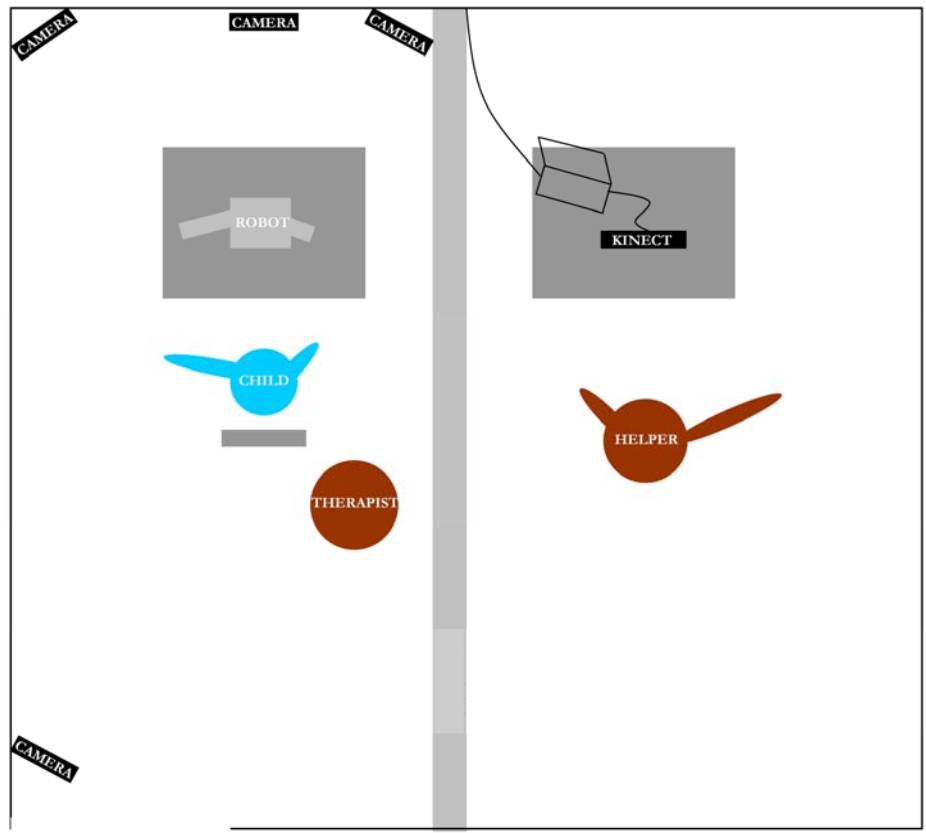

Fig. 2: Description of the experimental room

**5. Results/Discussion**

Both a qualitative and a quantitative analysis of the data were performed. The qualitative data suggest that some of the children understood that their movements were being mirrored by the robot Nao. Typical behaviors observed that are indicative of this are: the children were smiling contingently with the robot's mirroring movements, were performing motor acts and remained into the position until the robot imitated them, and switched attention between the experimenter and the robot, as the two agents were involved in a synchronous interaction. For the quantitative analysis of the data, due to the different duration of the trials, all the variables measured were normalized by time. Moreover, we mixed the data from the dyadic and triadic interaction procedures in the same pool of data. The Mann-Whitney test was used to analyze the difference between all the phases of the experiment. The results confirm only parts of the experimental hypothesis. Only Child#1 and Child#2 showed more attention and positive affect in the interaction with the Nao robot, suggesting the fact that the Nao robot has appealing characteristics for interaction. In the case of the initiations and imitations variable, only Child#1 and Child#3 show a higher frequency of these behaviors in the interaction with the Nao robot compared to the other phases of the study, therefore suggesting the superiority of the Nao procedure in facilitating imitations and initiations of motor acts. The Nao robot proved to be a better facilitator of shared attention only for Child#1. No significant results were found for free initiations.

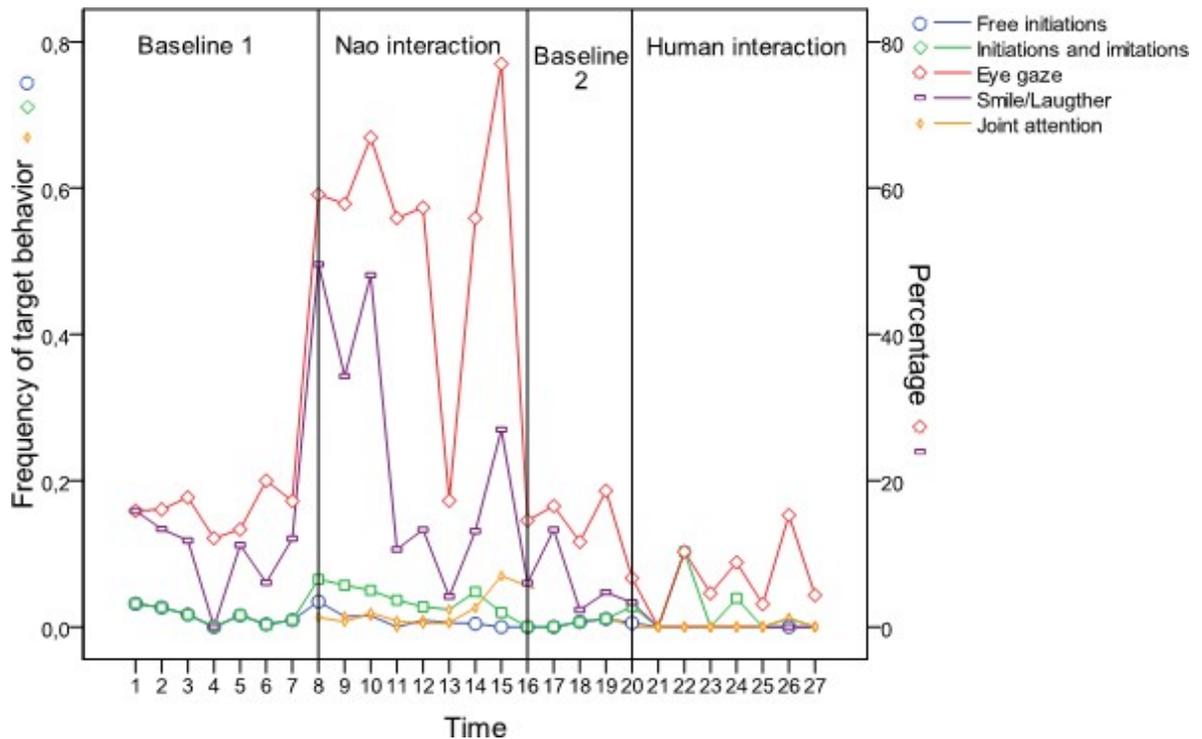

Fig. 3: The evolution of the outcome data for Child#1

## 5. Conclusion

Overall, the results of the study are mixed, showing some effect in some children, but not in all four of them. In a future study, we plan to repeat the procedure with a greater number of low-functioning autistic subjects in a clinical trial design. We consider that the major benefit of this study stands in the steps taken towards adapting standard single-subject paradigms into human-robot interaction. This kind of study represents a small step in advancing the engineering field towards the goal of actually providing a tool for clinicians and therapists.

**References:**

1. American Psychiatric Association (APA). (1994). Diagnostic and statistical manual of mental disorders (4th ed.).

2. Dawson, G., & Galpert, L. (1990). Mothers' use of imitative play for facilitating the social behavior of autistic children. *Development and Psychopathology, 2,* 151-162.

3. Meltzoff, A., & Gopnik, A. (1993). The role of imitation in understanding persons and developing a theory of mind. In S. Baron-Cohen, H. Tager-Flusberg & D.J. Cohen (Eds.), *Understanding other minds: Perspectives from autism* (pp. 335–366). Oxford: Oxford University Press.

4. Nadel-Brulfert, J. & Baudonniere, P. (1982). The social function of reciprocal imitation in 2-year-old peers. *International Journal of Behavioral Development*, 5:95–109

5. Nadel, J., Fontaine, A.M. (1989). Communicating by imitation: A developmental and comparative approach to transitory social competence. *In Social competence in developmental perspective* (Schneider BH, Attili G, Nadel J, Weissberg RP, editors). Dordrecht: Kluwer. pp 131-144.

6. Nadel, J. (2004). Do children with autism understand imitation as intentional interaction? *Cognitive and Behavioral Psychotherapies*, 4(2):165–177.

7. Stone, W., Ousley, O., & Little ford, C. (1997). Motor imitation in young children with autism: What's the object*?. Journal of Abnormal Child Psychology*, 25, 475–485.

8. Williams, J. H. G., Whiten, A. and Singh, T. (2004). A systematic review of action imitation in autistic spectrum disorder. *Journal of Autism and Developmental Disorders*, 34, 285-299.

9. A. Tapus, A. Peca, A. Aly, C. Pop, L. Jisa, S. Pintea, A. Rusu, and D. David, "Social Engagement of Children with Autism during Interaction with a Robot", *Proceedings of the 2nd International Conference on Innovative Research in Autism (IRIA)*, France, 2012.

10. A. Aly and A. Tapus, "Gestures Imitation with a Mobile Robot in the Context of Human Robot Interaction for Children with Autism", *Proceedings of the 3rd Workshop for Young Researchers on Human-Friendly robotics (HFR)*, Germany, 2010.

11. A. Aly, "Human Posture Recognition and Gesture Imitation with a Humanoid Robot", *Université Paris 6 (UPMC), Université de Sorbonne*, France, 2011.

12. A. Tapus, A. Peca, A. Aly, C. Pop, L. Jisa, S. Pintea, A. Rusu, and D. David, "Children with Autism Social Engagement in Interaction with Nao, an Imitative Robot - A Series of Single Case Experiments", *Interaction Studies*, 2012, 13 (3), pp.315-347.